\documentclass{article}

\PassOptionsToPackage{numbers}{natbib}

\usepackage{pbox}
\usepackage{booktabs}

\usepackage[preprint]{neurips_2019}
\usepackage{hyperref}

\usepackage{natbib}
\renewcommand{\bibname}{References}
\renewcommand{\bibsection}{\subsubsection*{\bibname}}

\usepackage{enumerate}

\usepackage{subfiles}

\usepackage{caption}
\usepackage{subcaption}
\usepackage{float}
\usepackage{tabularx}
\usepackage{wrapfig}

\usepackage{amsmath}
\usepackage{amssymb}
\usepackage{amsthm}
\usepackage{mathtools}
\usepackage{dsfont}

\usepackage[ruled,vlined]{algorithm2e}

\SetCommentSty{algcommstyle}

\usepackage{tikz}
\usetikzlibrary{bayesnet}

\usepackage{textcomp}
\usepackage{sourcecodepro}
\usepackage{listings}
\definecolor{commentgrey}{gray}{0.45}
\definecolor{backgray}{gray}{0.96}
\newcommand*{\cond}{\;\ifnum\currentgrouptype=16 \middle\fi|\;}

\newcommand*{\ttilde}{{\raise.17ex\hbox{$\scriptstyle\sim$}}}

\makeatletter
\newsavebox{\mybox}\newsavebox{\mysim}
\newcommand*{\distas}[1]{%
  \savebox{\mybox}{\hbox{\kern3pt$\scriptstyle#1$\kern3pt}}%
  \savebox{\mysim}{\hbox{$\sim$}}%
  \mathbin{\overset{#1}{\kern\z@\resizebox{\wd\mybox}{\ht\mysim}{$\sim$}}}%
}
\makeatother

\makeatletter
\def\moverlay{\mathpalette\mov@rlay}
\def\mov@rlay#1#2{\leavevmode\vtop{%
   \baselineskip\z@skip \lineskiplimit-\maxdimen
   \ialign{\hfil$\m@th#1##$\hfil\cr#2\crcr}}}
\newcommand*{\charfusion}[3][\mathord]{
  #1{\ifx#1\mathop\vphantom{#2}\fi\mathpalette\mov@rlay{#2\cr#3}}
  \ifx#1\mathop\expandafter\displaylimits\fi}
\makeatother

\newtheorem{theorem*}{Theorem}
\newtheorem{corollary*}[theorem*]{Corollary}
\newtheorem{proposition*}[theorem*]{Proposition}
\newtheorem{lemma*}[theorem*]{Lemma}

\theoremstyle{definition}

\newtheorem{definition*}{Definition}

\newtheoremstyle{algodesc}{}{}{}{}{\bfseries}{.}{ }{}%
\theoremstyle{algodesc}

\usepackage{color}

\begin{document}

\title{Active Federated Learning}
\author{
    Jack Goetz\\
    University of Michigan
    \And
    Kshitiz Malik\\
    Facebook Assistant
    \And
    Duc Bui\\
    University of Michigan
    \And 
    Seungwhan Moon\\
    Facebook Assistant
    \And
    Honglei Liu\\
    Facebook Assistant
    \And 
    Anuj Kumar\\
    Facebook Assistant
}
\date{\today}

\maketitle

\begin{abstract}
Federated Learning allows for population level models to be trained without centralizing client data by transmitting the global model to clients, calculating gradients locally, then averaging the gradients. Downloading models and uploading gradients uses the client's bandwidth, so minimizing these transmission costs is  important. The data on each client is highly variable, so the benefit of training on different clients may differ dramatically. To exploit this we propose \textit{Active Federated Learning}, where in each round clients are selected not uniformly at random, but with a probability conditioned on the current model and the data on the client to maximize efficiency. We propose a cheap, simple and intuitive sampling scheme which reduces the number of required training iterations by 20-70\% while maintaining the same model accuracy, and which mimics well known resampling techniques under certain conditions.
\end{abstract}

\section{Introduction}

% \begin{itemize}
%     \item Intro FL (example uses) + discuss main cost concerns (transmission costs + computation on client).
%     \item Foreshadow benefits of using loss and it's connections with other well known techniques.
%     \item State desire for Differential Privacy for each user in some situations, and briefly describe structure which can be exploited to provide said privacy with less disruption than pure naive Laplace noise. 
% \end{itemize}

As machine learning models are deployed in the real world, the assumptions under which they were developed are often shown to be incompatible with user requirements. One assumptions is unrestricted access to the training data, either on a single machine or distributed over many researcher controlled machines. Due to privacy concerns users may not want to transmit data from their personal devices, making such centralized training impossible. Federated Learning enables the training of models on this data, but transmission costs between the server and the client are high, and reducing these costs is important. In this paper we introduce \textit{Active Federated Learning} (AFL) to preferentially train on users which are more beneficial to the model during that training iteration. Motivated by ideas from Active Learning, we propose using a value function which can be evaluated on the user's device and returns a valuation to the server indicating the likely utility of training on that user. The server collects these valuations and converts them to probabilities with which the next cohort of users is selected for training. By using simple a value function related to the loss the user's data suffers under the current model, we can reduce the number of training rounds required for the model to achieve a specified level of accuracy by 20-70\%.

\section{Related Work}

% \begin{itemize}
%     \item FL related
%     \item resampling/AL related (just a little)
%     \item Summarize differences with both classic resampling and active learning implying the need for new algorithm.
%     \item Diff Priv related
% \end{itemize}

Since its introduction \citep{mcmahan2016communication, yang2019federated}, reducing the communication costs of Federated Learning has been an important goal \citep{konevcny2016federated, caldas2018expanding}. However as discussed in \citet{li2019federated} there are few existing techniques which change the method of selecting users. In \citet{hartmann2018federated} the author suggests stratification based on contextual information about the users, and in \citet{nishio2019client} the authors group users based on hardware characteristics. In contrast our work is closer to Active Learning (AL) \citep{settles2009active} where the selection policy is dependant on the current state of the model and the data on each user. In both paradigms training data must be selected under imperfect information; in AL the covariates are fully known, but the label of candidate data points is unknown, whereas in AFL both labels and covariates are fully known on each client, but only a summary is returned to the server. Additionally, in standard AL individual data points may be selected in an unconstrained manner, whereas in AFL we train on all data points on each selected user, creating predetermined subsets of data.

\section{Background and Notation}

Assume we have labelled data $(x,y)$ and a model for predicting $y \in \mathcal{Y}$ given $x \in \mathcal{X}$ which we denote by $\hat{y} = f(x ; \mathbf{w})$, where $\mathbf{w} \in \mathbb{R}^d$ are our model parameters. These model parameters will be learned by minimizing some loss function $l(x, y ; \mathbf{w})$. Assume our training data is distributed over multiple clients (or users) $\mathcal{U} = \{U_1,...,U_K\}$, where we denote the data of client $U_k$ by $(\mathbf{x}_k, \mathbf{y}_k) \in \mathcal{X}^{n_k} \times \mathcal{Y}^{n_k}$. Our model parameters will be learned during training iterations, so we will let $\mathbf{w}^{(t)}$ denote the value of our parameters at training iteration $t$. During each training iteration we select a subset of users $\mathcal{S}^{(t)} \subset \mathcal{U}, |\mathcal{S}^{(t)}| = m$ and send $\mathbf{w}^{(t)}$ to each user in the set. Each user then performs some training $\mathcal{T}$ using their local data and produce updated model parameter values $\mathbf{w}_k^{(t+1)} = \mathcal{T}(\mathbf{x}_k, \mathbf{y}_k ; \mathbf{w}^{(t)})$. In its most simple form this training could be a single step of gradient descent, though in practice it is often more complicated, such as multiple passes of SGD. These updated model parameter values are then returned to the server and aggregated to produce the next model parameters using Federated ADAM \cite{leroy2019federated}. In traditional Federated Learning the subsets $\mathcal{S}^{(t)}$ are selected uniformly at random and independently at each iteration. Our goal in AFL is to select our subsets $\mathcal{S}^{(t)}$ such that fewer training iterations are required to obtain a good model.

\section{Active Federated Learning (AFL)}

% \begin{itemize}
%     % \item Give generic AFL where value() is quite generic. Discuss good properties that the value function should have.  
%     \item Give loss based value function and show similarity to subsampling when you have large (and non-separated) class imbalance, similarity to uncertainty sampling from AL (i.e. you select users that have hard data points), similarity to sampling higher volume users when massive data count imbalance. If the noise depends on distance from the margin as in (5) in \cite{blaschzyk2018improved} going to replicate margin based subsampling. For class imbalance and volume of data, calculate correlation between loss values and other values. Probably quite correlated. Point out that it does not need prior knowledge of which structure is being exploited ahead of time, and when reported does not reveal which of these structures user has (so can exploit these "privately" even without Differential Privacy ideas). Also is able to use both structures at the same time (and by picking averaging power can balance them automatically).
%     % \item Discuss initialization of AFL metrics. They do require every user to get an updated version of the model and do a single forward pass over their data. Computationally this  is equivalent to a single update on each user, but AFL saves much more than a single Epochs worth of computation.
%     % \item Discuss limitations of having to select batches of users as opposed to individual data points. 
% \end{itemize}

Inspired by the structure of classical AL methods, we propose the AFL framework which aims to select an optimized subset of users based on a \textit{value} function that reflects how useful the data on that user is during each training round. Formally, we define a function $\mathcal{V}: \mathcal{X}^{n_k} \times \mathcal{Y}^{n_k} \times \mathbb{R}^d \rightarrow \mathbb{R}$ which is evaluated on each user. Once evaluated, each user $U_k$ returns a corresponding \textit{valuation} $v_k \in \mathbb{R}$ to the server, which is used to calculate the sampling distribution for the next training iteration. The valuations are a function of $\mathbf{w}^{(t)}$, but since transmitting the model is expensive we only get fresh valuations of users during an iteration in which we train on them, meaning that

\vspace{-8pt}

\begin{equation*}
\begin{aligned}
    v_k^{(t+1)} = 
    \begin{cases}
        \mathcal{V}(\mathbf{x}_k, \mathbf{y}_k ; \mathbf{w}^{(t)}) & \text{ if } U_k \in S_t\\
        v_k^{(t)} & \text{ otherwise.}
    \end{cases}
\end{aligned}
\end{equation*}

\vspace{-5pt}

Ideally the computation of the value function should require minimal additional computation, since the computations are done using the clients hardware, and should not reveal too much about the data on each client. Once the server has all valuations it converts them into a sampling distribution. 

\vspace{-1pt}

\begin{figure}[h!]
	\centering
	\begin{tabular}{cc}
		\includegraphics[width=0.8\textwidth,height=0.19\textheight,clip = true]{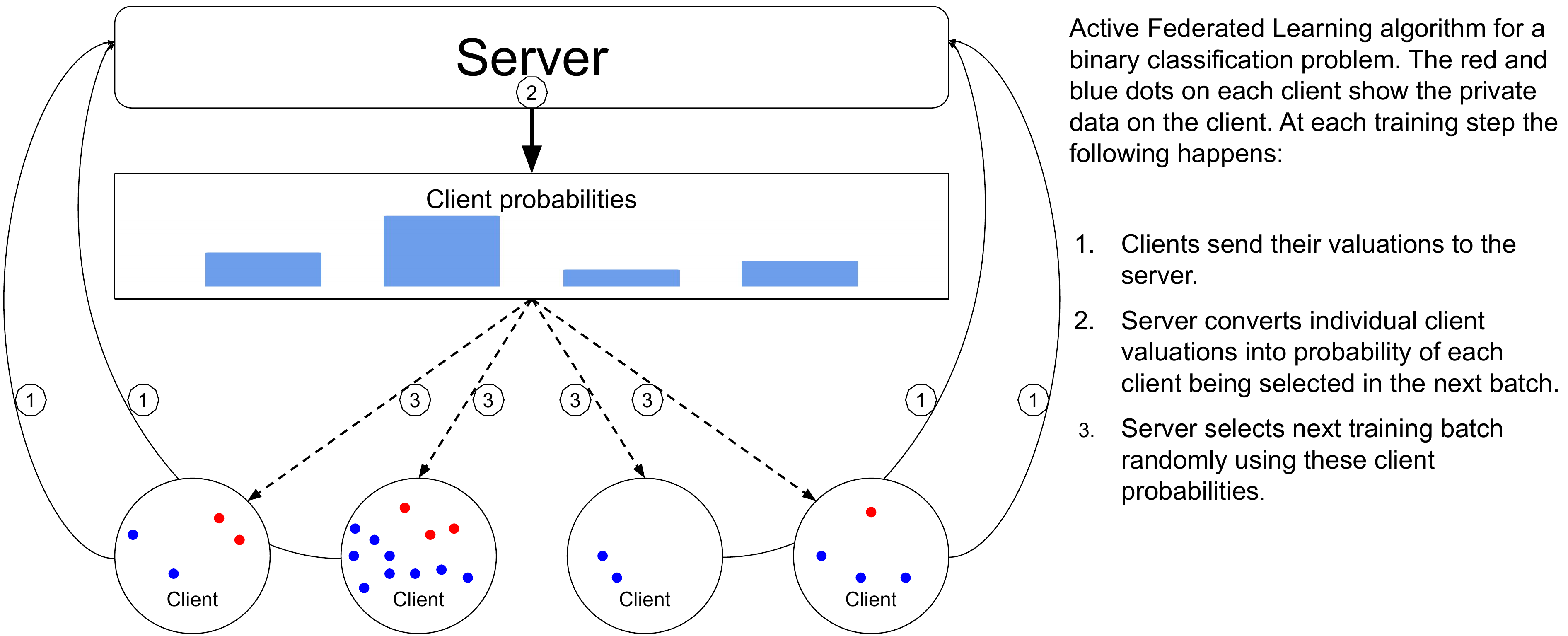}\\
	\end{tabular}
	\label{fig:alf_setup}
	\caption{\textbf{Active Federated Learning framework} for a binary classification problem.}
\end{figure}

% \vspace{-1.5em}

\subsection{Loss valuation}

One very natural value function is to use the loss of the users data $v_k = \frac{1}{\sqrt{n_k}} l(\mathbf{x}_k, \mathbf{y}_k ; \mathbf{w})$. It is already calculated during model training and is increasing with how poorly the model performs on the clients data. Additionally it mimics common resampling techniques when the required structure is present in the data. If there is extreme class imbalance and weak separation of the classes, data points of the minority class will have significantly higher loss than majority class data points. Therefore we will prefer users with more minority data, mimicking resampling the minority class data. Similarly if the noise depends on the distance from the classification boundary such as in \citep{blaschzyk2018improved}, using the loss replicates margin based resampling techniques. Finally if all data points are equally valuable then users with more data will be given higher valuations. Most importantly these adaptations to the data do not require the practitioner to know the specific structure being exploited. This is particularly important in the Federated setting, where information about the data is limited. 

\subsection{Differential Privacy}

Even summarizing the client data with a single float may reveal too much information. To properly protect users the value function should be reported using a Differentially Private mechanism \cite{dwork2014algorithmic}. The noise introduced to maintain Differential Privacy may mislead the server into selecting sub-optimal clients. However there is structure which might be exploited to reduce the corruption while still maintaining privacy. One is that many value functions, such as the loss, are not expected to change dramatically within a small number of training rounds. Thus we may be able to query whether a valuation has changed dramatically before querying the new value, similar to the Sparse Vector technique, to reduce the number of queries. We may also be able to adapt our value function to be more amenable to Differential Privacy. For example the loss value function has unbounded sensitivity and requires clipping to provide Differential Privacy. However returning a count of high loss data points has sensitivity $1$ and may be less affected by the privacy providing noise. Adding privacy guarantees is an important challenge in AFL and is the subject of much future work.

\section{Experimental Results}

%\shane{Can we add descriptions of FB and Reddit datasets here - for Comms review purposes e.g. where we got the data from, what kinds of data, etc.}

%We compared AFL to the standard uniform selection on two datasets; one on the Reddit dataset, the other on an Internal Message dataset. For the Reddit dataset we predicted the binary label 'controversially' based on the comment text, and selected 8K users at random from the November 2017 data set, similar to \citet{bagdasaryan2018backdoor} but we only excluded users with +100K messages. We removed comments being responded to from the messages, and empty messages. The Reddit dataset has many users who post few comments, but a long tail of power users. The Internal Messages data set has messages from a popular messaging app and the task was to predict whether the message was replied to using a sticker. The messages were collected, de-identified and annotated automatically (no human reading or labelling of messages).

We compared AFL to the standard uniform selection on two datasets; one on the Reddit dataset, the other on the Sticker Intent dataset. The Reddit dataset is a publicly available~\cite{reddit_comments_dump} dataset consisting of comments from users on \emph{reddit.com}. The authors were not involved in collecting this dataset. For the Reddit dataset we predicted the binary label 'controversially' based on the comment text, and selected 8K users at random from the November 2017 data set, similar to \citet{bagdasaryan2018backdoor} but only excluding users with +100K messages. We removed comments being responded to from the messages, and empty messages. The Reddit dataset has many users who post few comments, but a long tail of power users. The Sticker Intent dataset has randomly selected, anonymized messages from a popular messaging app. The task was binary classification - predict whether a message was replied to using a sticker. Messages in this set were collected, de-identified, and annotated automatically; the messages were not read or labeled by human annotators.

\begin{algorithm}[t]
\caption{Sampling algorithm} \label{alg:sampling}
  \SetAlgoLined
  \KwIn{Client Valuations $\{v_1,...,v_K\}$, tuning parameters $\alpha_1,...,\alpha_3$, number of clients per round $m$}
  \KwOut{Client indices $\{k_1,...,k_m\}$}
  Sort users by $v_k$\\
  For the $\alpha_1 K$ users with smallest $v_k$, $v_k = -\infty$ \\
  \For{$k$ from $1$ to $K$}{
    $p_k \propto e^{\alpha_2 v_k}$\\
  }
  Sample $(1 - \alpha_3)m$ users according to their $p_k$, producing set $\mathcal{S}'$ \\
  Sample $\alpha_3m$ from the remaining users uniformly at random, producing set $\mathcal{S}''$\\
  return $\mathcal{S} = \mathcal{S}' \cup \mathcal{S}''$\\
\end{algorithm}
% \vspace{-2em}

\begin{table}[h!]
    \caption{Reddit dataset statistics}
    \begin{tabular}{c|cccccccc}
    \toprule
        & messages & users & \% label $1$ & mean messages/user & median messages/user\\
        \hline
        Train & 124638 & 7527 & 0.021 & 16.6 & 3\\
        Test & 15568 & 3440 & 0.021 & 4.5 & 2\\
    \bottomrule
    \end{tabular}
    \label{tab:reddit_dataset}
\end{table}

Algorithm \ref{alg:sampling} for converting the valuations into a sampling distribution has 3 tuning parameters: The $\alpha_1$ proportion of users with the smallest valuations will have their valuations set to $-\infty$. They can still be selected by random sampling. $\alpha_2$ is our softmax temperature. $\alpha_3$ is the proportion of users which are selected uniformly at random. In our experiments we used $\alpha_1 = 0.75, \alpha_2 = 0.01, \alpha_3 = 0.1$. We chose $\alpha_2$ to ensure that the softmax did not produce $p_k = 0$ from underflow errors, and $\alpha_1, \alpha_3$ were both chosen based on initial experiments on Sticker Intent dataset. The underlying model trained with Federated Learning used a 64 dimensional character level embedding, a 32 dimensional BLSTM, and an MLP with one 64 dimensional hidden layer. The number of users in each Federated round was 200, and on each user 2 passes of SGD was performed with a batch size of 128. The learning rates for both local SGD and Federated ADAM were tuned separately for Random Sampling and AFL and the optimal learning rates were used for each.

\begin{figure}[h!]
	\centering
	\begin{tabular}{cc}
		\includegraphics[width=0.5\textwidth,height=0.17\textheight,clip = true]{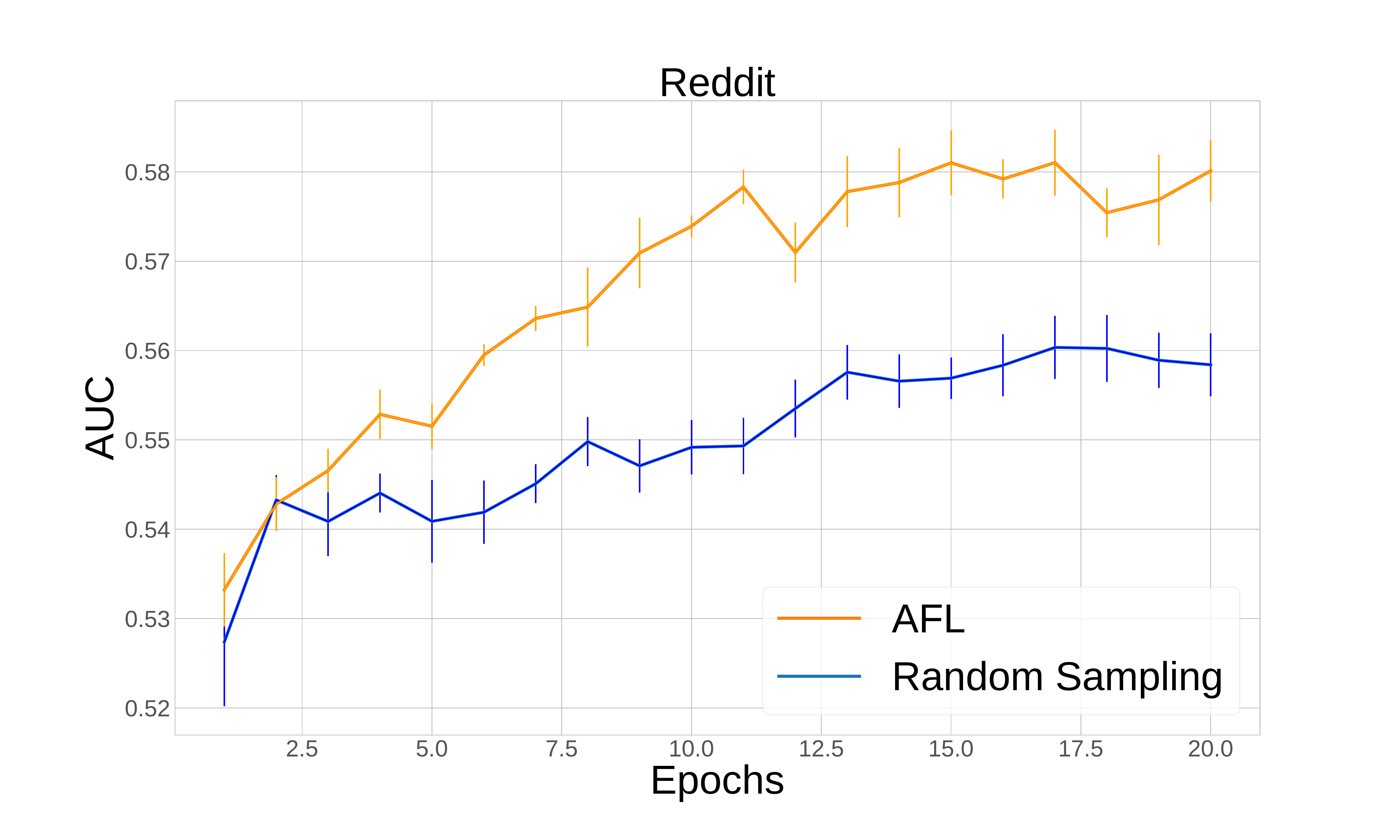}
		\includegraphics[width=0.5\textwidth,height=0.17\textheight,clip = true]{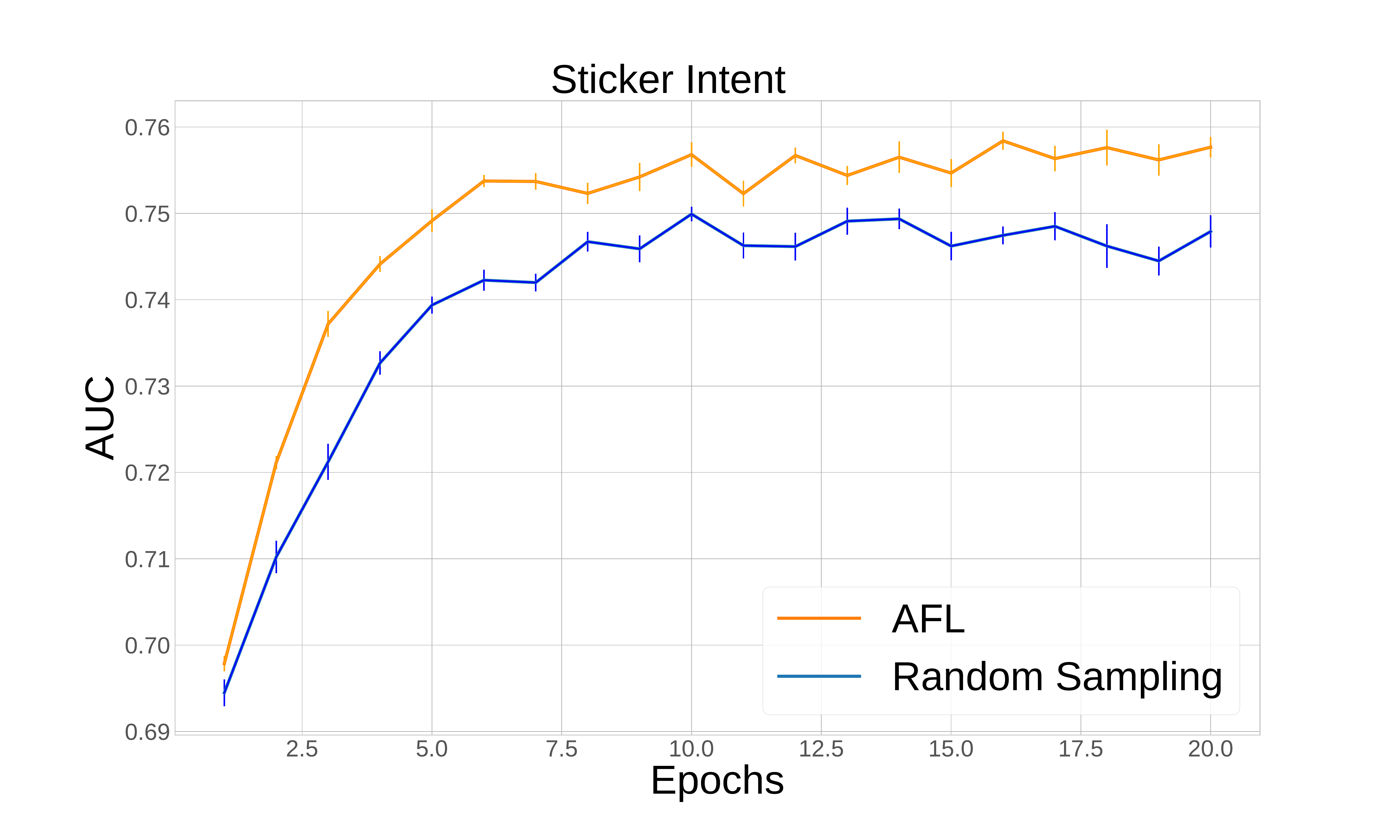}\\
	\end{tabular}
	\caption{Comparison of AUC increase on Reddit and Sticker Intent datasets}	
	\label{fig:afl}
\end{figure}

Figure \ref{fig:afl} shows the AUC after each Epoch under uniform random selection of users, and with AFL selection, showing mean and standard errors from 10 repetitions on test data. AFL trains models of the same performance using 20-70\% fewer Epochs (where one Epoch is enough training rounds to train on each client once in expectation under random sampling). 

\vspace{-2pt}

% \vspace{-10pt}

% \subsection{Comparison with Resampling of minority class}

One difference between AFL and server-side resampling techniques is that AFL selects data points by user, whereas server-side resampling can select arbitrary subsets. To explore the significance of this restriction we compared the gains from oversampling of label $1$ data \cite{he2008learning} and server-side learning against AFL using the value function $v_k = \sum \mathbf{1}_{y_{i,k} = 1}$ and Federated training, using the Reddit dataset. The level of resampling and learning rates were tuned for server training, as were the temperature $\alpha_2$ and the learning rates for Federated training, and all other tuning parameters were kept the same. Our results suggest that there is significant loss from selecting users, as the difference between Random Sampling and Active Sampling is much larger for server-side learning.

\vspace{-0.5em}

\begin{table}[h!]
    \centering
    \begin{tabular}{c|cc}
    \toprule
        & Random Sampling & Active Sampling \\
        \hline
        Server selection of data points & 0.559 & 0.615\\
        Federated selection of clients & 0.552 & 0.578\\
    \bottomrule
    \end{tabular}
    \label{tab:resampling}
\end{table}

\vspace{-1em}

\section{Conclusion and Further directions}

% \begin{itemize}
%     \item Changing live data on each user, as new data comes in and old data becomes stale and is discarded. Use bandit style algorithm to ensure users are checked sufficiently often. Suggest connections to restless bandits. Bandit weighting likely needed if class balance, data point number, difficulty evenly spread, but in those cases benefit from AFL likely to be small anyway.
%     \item Users are not 100\% reliable in returning values. Use ideas from cost based Active Learning, such as have each user cost = P(successfully return gradient).
%     \item Possibility of not using all data points on each user to train, or having resampling on user as well.
%     \item Understand use on non-binary classification problems.
% \end{itemize}

In this paper we proposed Active Federated Learning (AFL), the first user cohort selection technique for FL which actively adapts to the state of the model and the data on each client. This adaptation allows us to train models with 20-70\% fewer iterations for the same performance. Giving formal privacy guarantees is vital future work, but there are many other interesting extensions as well. These experiments were done under simplifying conditions which do not take into account many problems Federated Learning faces in practice, and which AFL may be able to help alleviate. For example clients may have different rates of availability for training. This availability may be correlated with the data on the client, resulting in bias in our model if not corrected. AFL which also takes reliability into account may be used to reduce this bias by increasing the rate at which we try to train on unreliable users. Another challenge is that clients are constantly gathering (and potentially forgetting) data, and in many cases the distribution may be non-stationary. Maintaining the benefits of AFL may require a principled way of ensuring no user goes too long without having their valuation refreshed. Finally our experiments and analyses focused on the classification setting, but the loss value function can be used for any supervised problem, and understanding AFL with more complex models would be an interesting research direction.

\bibliographystyle{apa}
\bibliography{workshop.bib}

% \clearpage

% \section{Appendix}

% \subfile{sections/appendix}

%\begin{figure}[h!]
	%\caption{}
	%\centering
	%\begin{tabular}{cc}
		%\includegraphics[width=0.99\textwidth,height=0.25\textheight,clip = true]{}\\
	%\end{tabular}
	%\label{fig:diff}
%\end{figure}
%
%
%
%\begin{equation*}
%\begin{aligned}
%\end{aligned}
%\end{equation*}
%
%
%
%\begin{lstlisting}[language=Python,basicstyle=\tiny]
%
%
%
%
%\end{lstlisting}

% \vfill
% \FloatBarrier \pagebreak
% \nocite{*}

\end{document}

\subsection{Converting values into probabilities}

% Our algorithm for converting the valuations into a sampling distribution has 3 tuning parameters: The $\alpha_1$ proportion of users with the smallest valuations will have their valuations set to $-\infty$. They can still be selected by random sampling. $\alpha_2$ is our (inverse) softmax temperature. $\alpha_3$ is the proportion of users which are selected uniformly at random. 

% \begin{algorithm}[h]
% \caption{Sampling algorithm} \label{alg:sampling}
%   \SetAlgoLined
%   \KwIn{Client Valuations $\{v_1,...,v_K\}$, tuning parameters $\alpha_1,...,\alpha_3$, number of clients per round $m$}
%   \KwOut{Client indices $\{k_1,...,k_m\}$}
%   Sort users by $v_k$\\
%   For $\alpha_1 K$ users with smallest $v_k$, $v_k = -\infty$ \\
%   \For{$k$ from $1$ to $K$}{
%     $p_k \propto e^{\alpha_2 v_k}$\\
%   }
%   Sample $(1 - \alpha_3)m$ users according to their $p_k$ giving set $\mathcal{S}'$ \\
%   Sample $\alpha_3m$ from the remaining users uniformly at random giving set $\mathcal{S}''$\\
%   return $\mathcal{S} = \mathcal{S}' \cup \mathcal{S}''$\\
% \end{algorithm}

% In our experiments we used $\alpha_1 = 0.75, \alpha_2 = 0.01, \alpha_3 = 0.1$. $\alpha_2$ was chosen to ensure that the softmax only gave users with $v_k = -\infty$ zero probability of being selected, and $\alpha_1, \alpha_3$ were both chosen based on experiments on the FB dataset, but were not carefully tuned on either dataset. 

\subsection{Data set details}

\subsection{Network architecture}